\documentclass{article} 
\usepackage{iclr2027_conference,times}

\usepackage{amsmath,amsfonts,bm}

\def\eqref#1{equation~\ref{#1}}

\def\1{\bm{1}}

\DeclareMathAlphabet{\mathsfit}{\encodingdefault}{\sfdefault}{m}{sl}
\SetMathAlphabet{\mathsfit}{bold}{\encodingdefault}{\sfdefault}{bx}{n}

\usepackage[utf8]{inputenc} 
\usepackage[T1]{fontenc}    
\usepackage[hidelinks]{hyperref}  
\usepackage{url}            
\usepackage{booktabs}       
\usepackage{amsfonts}       
\usepackage{nicefrac}       
\usepackage{microtype}      
\usepackage{xcolor}         
\usepackage{graphicx}

\usepackage{amsmath}
\usepackage{amssymb}
\usepackage{mathtools}
\usepackage{amsthm}
\usepackage{makecell}
\usepackage{comment}
\usepackage[capitalize,noabbrev]{cleveref}

\usepackage{subcaption}
\usepackage{lipsum}
\usepackage{algorithm}
\usepackage{algpseudocode}
\usepackage{algorithmicx}
\usepackage{bm}
\usepackage{wrapfig}
\usepackage[super]{nth}
\usepackage{pifont}
\usepackage{bbding}

\title{Prior-Fitted and Frozen Scorers for Multimodal Contrastive Learning}

\author{
  \textbf{Tillmann Rheude\textsuperscript{1,2$^\dagger$}}, 
  \textbf{Roland Eils\textsuperscript{1,2,3$^\dagger$}}, 
  \textbf{Benjamin Wild\textsuperscript{1$^\dagger$}}\\
  \textsuperscript{1}Berlin Institute of Health, Charité - Universitätsmedizin Berlin, 
  \textsuperscript{2}Department of Mathematics\\and Computer Science, Freie Universität Berlin, 
  \textsuperscript{3}Intelligent Medicine Institute, Fudan University\\
  $^\dagger$\texttt{\{tillmann.rheude, roland.eils, benjamin.wild\}@bih-charite.de}
}

\iclrfinalcopy 

\usepackage[
  frozencache,
  cachedir=minted-cache
]{minted}
\setminted{
    style=borland,
    linenos,
    numbersep=4pt,
    xleftmargin=1em,
}

\usepackage[acronym,toc,shortcuts]{glossaries}
\glsdisablehyper
\newacronym{ukb}{UKB}{UK Biobank}
\newacronym{sota}{sota}{state-of-the-art}
\newacronym{flops}{FLOPS}{Floating Point Operations}
\newacronym{ce}{CE}{Cross Entropy}
\newacronym{bce}{BCE}{Binary Cross Entropy}
\newacronym{ecg}{ECG}{electrocardiogram}
\newacronym{ehr}{EHR}{electronic health record}
\newacronym{mri}{MRI}{Magnetic Resonance Images}
\newacronym{ct}{CTPA}{computed tomography pulmonary angiography}
\newacronym{prs}{PRS}{polygenic risk score}
\newacronym{auroc}{AUROC}{area under the receiver operating characteristic}
\newacronym{vit}{ViT}{Vision Transformer}
\newacronym{mlp}{MLP}{Multi-Layer Perceptron}
\newacronym{cnn}{CNN}{Convolutional Neural Network}
\newacronym{sd}{SD}{standard deviation}
\newacronym{se}{SE}{standard error}
\newacronym{gpu}{GPU}{Graphics Processing Unit}
\newacronym{mip}{MIP}{multilinear inner product}
\newacronym{tc}{TC}{total correlation}
\newacronym{moe}{MoE}{mixture-of-experts model}
\newacronym{poe}{PoE}{product-of-experts model}
\newacronym{ood}{OOD}{Out-of-Distribution}
\newacronym{cam}{CAM}{class activation map}
\newacronym{gpe}{GPE}{geometry-preserving encoder}
\newacronym{hpc}{HPC}{high-performance cluster}
\newacronym{icu}{ICU}{intensive care unit}
\newacronym{jvp}{JVP}{Jacobian-vector product}
\newacronym{vlm}{VLM}{vision-language model}

\newacronym{scm}{SCM}{structural causal model}
\newacronym{pfs}{PFS}{Prior-Fitted Scorer}
\newacronym{pfn}{PFN}{Prior-Data Fitted Network}
\newacronym{dag}{DAG}{directed acyclic graph}
\newacronym{pe}{PE}{pulmonary embolism}

\newcommand{\ie}{\textit{, i.e., }}
\newcommand{\eg}{\textit{, e.g., }}

\begin{document}

\maketitle

\begin{abstract}
    Multimodal contrastive learning typically relies on manually specified scoring functions, such as cosine similarity, to quantify the similarity between representations from different modalities. 
    Several such functions have been recently proposed to extend contrastive learning beyond two modalities.
    In principle, however, any function that assigns high compatibility to matched samples and low compatibility to mismatched ones could serve as a scorer.
    Motivated by this broader view, we instantiate the scorer with a Transformer and find that even a randomly initialized model, when kept frozen, can outperform strong handcrafted multimodal contrastive baselines, whereas optimizing the scorer jointly with the encoders is less effective.    
    Building on this observation, we introduce the \ac{pfs}: a Transformer scorer pre-trained on synthetic multimodal data generated from \ac{scm} priors before transferring it to real-world data.
    The SCM exposes the scorer to diverse dependency structures, including shared, pairwise, and private information, providing a structured prior over higher-order cross-modal similarity.
    Across four real-world datasets, frozen Transformer scorers are competitive to contrastive baselines, and prior-fitting further improves the random initialization.
    More broadly, our results suggest that multimodal contrastive learning need not rely on manual similarity functions: Transformers can provide strong inductive biases, while causal pretraining can further shape these biases.\footnote{The code repository including a pretrained \acs{pfs} checkpoint is available on \href{https://github.com/TillmannRheude/prior_fitted_scorer}{GitHub}.}
\end{abstract}

\section{Introduction}
\label{sec:introduction}

Contrastive learning has become a central method for learning transferable multimodal representations. A famous example is CLIP \citep{Radford_Kim_Hallacy_Ramesh_Goh_Agarwal_Sastry_Askell_Mishkin_Clark_etal._2021}, which aligns image and text embeddings by contrasting matched and mismatched pairs using cosine similarity as a scoring function. However, many real-world problems involve more than two modalities, motivating objectives that explicitly capture higher-order dependencies. 
Symile \citep{Saporta_Puli_Goldstein_Ranganath_2024} introduces the \ac{mip} extending the standard dot product to capture higher-order multimodal interactions.
Other approaches, such as Gram \citep{Cicchetti_Grassucci_Sigillo_Comminiello_2025} and Triangle \citep{,Cicchetti_Grassucci_Comminiello_2025}, replace pairwise cosine similarity with geometry-based measures that jointly score multiple modality embeddings. 
These methods differ in how multimodal similarity is defined, but they share a common design choice: the scoring function is manually defined. This raises a more fundamental question: if similarity can, in principle, be expressed by many different functions, why restrict the scorer to a particular hand-designed manual form rather than represent and learn it more generally?

Neural networks provide a natural way to instantiate such a general scoring function. Particularly, a Transformer \citep{Vaswani_Shazeer_Parmar_Uszkoreit_Jones_Gomez_Kaiser_Polosukhin_2017} is a convenient realization of this idea, since modality embeddings can be represented as tokens and processed jointly by the same scorer across different modality counts. 
This avoids designing a new manual compatibility function for each setting and also allows missing modalities to be handled, which is particularly attractive in domains such as medicine \citep{Acosta_Falcone_Rajpurkar_Topol_2022}. Using a Transformer scorer introduces additional computation compared with closed-form similarities, but this cost is limited by the small number of modality tokens and by keeping the scorer frozen during downstream training. Even without any task-specific training, a randomly initialized and frozen Transformer already defines a nontrivial nonlinear function over its inputs.
We therefore investigate whether such Transformer-based functions can serve as effective multimodal scoring rules.
Surprisingly, we find that a randomly initialized Transformer already provides a strong multimodal scorer when kept frozen during training. In this case, only the modality encoders are trained. 

\begin{figure}[t]
    \centering
    \includegraphics[width=1.0\linewidth]{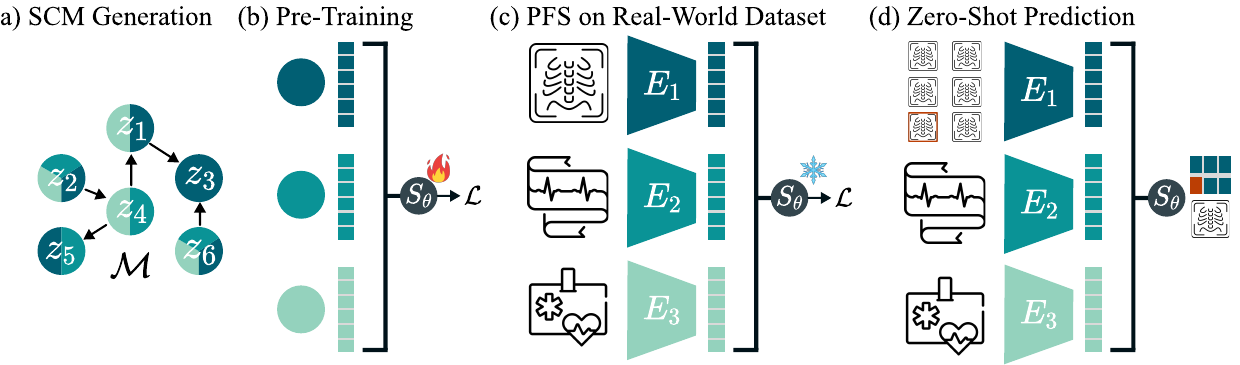}
    \caption{
        Overview of the \acf{pfs}, illustrated on the trimodal MIMIC-Symile setting \citep{Saporta_Puli_Goldstein_Ranganath_2024}. 
        \textbf{(a)} We sample latent causal structures and generate synthetic multimodal tuples from an \ac{scm} prior $\mathcal{M}$.
        \textbf{(b)} A Transformer scoring function $s_\theta$ is pretrained on these synthetic tuples to distinguish compatible from mismatched multimodal combinations with loss $\mathcal{L}$, thereby shaping its inductive bias.
        \textbf{(c)} The \ac{pfs} is transferred to the real dataset and kept frozen while the modality encoders $E_i$ are optimized through the contrastive objective.
        \textbf{(d)} After extensive tuning, the \ac{pfs} is evaluated on zero-shot predicting the future chest X-ray. 
        Overall, a \ac{pfs} replaces a manually specified similarity function with a Transformer whose structure is learned from a synthetic causal prior.
    }
    \label{fig:overview}
\end{figure}

Random initialization, however, leaves the scoring function entirely unguided. We therefore ask whether its inductive bias can be shaped before exposure to real data.
Inspired by \acp{pfn} \citep{Müller_Hollmann_Pineda-Arango_Grabocka_Hutter_2022}, which learn inference procedures by training Transformers across synthetic tasks sampled from a prior, we pretrain the scorer on synthetic multimodal embeddings generated from \acp{scm} (\Cref{fig:overview}).
Unlike meta-learning approaches that pretrain a model to perform in-context learning, we pretrain only the multimodal scorer. The modality-specific encoders are learned from scratch for each dataset because the underlying inputs can differ\eg images, time series, text, or tabular measurements.
Related work has demonstrated that synthetic task distributions can support effective pretraining across a wide range of problems, including tabular prediction \citep{Hollmann_Müller_Purucker_Krishnakumar_Körfer_Hoo_Schirrmeister_Hutter_2025}, Bayesian optimization \citep{Müller_Feurer_Hollmann_Hutter_2023}, clustering \citep{Zhao_Wang_Tan_Zhang_2026}, survival analysis \citep{Böhm_Purucker_Hutter_Schlosser_2026}, uncertainty prediction \citep{Shi_Liu_Yu_Lu_Elias_Cheng_Kou_Chen_Liu_Huang_etal_2026}, and causal inference \citep{Robertson_Reuter_Guo_Hollmann_Hutter_Schölkopf_2025,Ma_Frauen_Javurek_Feuerriegel_2026}.
Here, we use \acp{scm} for a different purpose: to shape the inductive bias of a multimodal contrastive scorer before transferring it to real-world data.
Concretely, our \ac{scm} prior generates latent causal variables and maps them into modalities with shared, pairwise, and modality-specific information. We use the synthetic modalities to pretrain a Transformer as a contrastive scorer and transfer the resulting \ac{pfs} to real-world trimodal datasets.

Across four datasets, frozen randomly initialized Transformer scorers are competitive with well-tuned baselines trained by manual scoring functions, while \ac{scm}-based prior-fitting consistently improves upon random initialization. These results indicate that the scorer architecture itself contains an important inductive bias, and that this bias can be shaped through synthetic pretraining. Together, these findings broaden the design space of multimodal contrastive learning from choosing among manual similarity functions to designing, learning, and prior-fitting expressive scoring architectures.

Our main contributions are:
\begin{itemize}
    \item We show that a frozen Transformer scorer can serve as a strong competitor to manual  multimodal similarity functions despite having randomly initialized parameters.
    \item We introduce the \acf{pfs}, which pretrains the frozen Transformer scorer on synthetic \ac{scm}-generated multimodal data in order to shape its inductive bias toward higher-order multimodal similarity.
    \item We demonstrate that \ac{scm}-based prior-fitting further improves the already strong frozen Transformer scorer across real-world multimodal retrieval tasks.
\end{itemize}
\section{Related Work}
\label{sec:related_work}

\paragraph{Multimodal Contrastive Learning}
Multimodal contrastive learning has largely relied on manually designed similarity functions. Vision-language models such as CLIP \citep{Radford_Kim_Hallacy_Ramesh_Goh_Agarwal_Sastry_Askell_Mishkin_Clark_etal._2021} and ALIGN \citep{Jia_Yang_Xia_Chen_Parekh_Pham_Le_Sung_Li_Duerig_2021} use pairwise similarity, with later approaches such as LiT \citep{Zhai_Wang_Mustafa_Steiner_Keysers_Kolesnikov_Beyer_2022} and SigLIP \citep{Zhai_Mustafa_Kolesnikov_Beyer_2023} refining the corresponding training objectives. Extensions to more than two modalities introduce increasingly specialized constructions, including Symile \citep{Saporta_Puli_Goldstein_Ranganath_2024}, Triangle \citep{Cicchetti_Grassucci_Comminiello_2025}, CoMM \citep{Dufumier_Navarro_Tuia_Thiran_2025}, ConFu \citep{Koutoupis_Zervou_Kontras_DeVos_Tsakalides_Tsagkatakis_2026}, and Gram \citep{Cicchetti_Grassucci_Sigillo_Comminiello_2025}, whose scoring rules impose specific assumptions on how modalities interact and can depend on the modality configuration. Learned similarity functions have also been explored more directly\eg by replacing fixed inner-product similarity with a learned parametric measure \citep{Liu_Liu_Rehg_Song_2019}, and learning an MLP-based similarity function between representations within a self-supervised objective \citep{Zhang_Qiu_Ke_Süsstrunk_Salzmann_2022}. Learned critics provide an alternative case: Deep InfoMax \citep{Hjelm_Fedorov_Lavoie_Marchildon_Grewal_Bachman_Trischler_Bengio_2019} and CPC \citep{Oord_Li_Vinyals_2018} parameterize the contrastive scoring function rather than fixing it analytically. Our finding that even frozen random Transformers can be effective is also related to work on random features and untrained neural networks, where fixed random nonlinear mappings can preserve useful structure \citep{Rahimi_Recht_2007,Giryes_Sapiro_Bronstein_2016,Amid_Anil_Kotłowski_Warmuth_2022}. We build on these ideas by using a Transformer as a flexible multimodal scorer that can operate across varying modality combinations without redesigning the scoring rule.

\paragraph{Causal Models for Representation Learning}
\Acfp{scm} describe data-generating processes through a \ac{dag} and structural assignments that map parent variables and exogenous noise to their descendants \citep{Peters_Janzing_Schölkopf_2017}. Causal structure has increasingly been connected to contrastive learning\eg by interpreting augmentations as interventions and enforcing invariance across causal mechanisms \citep{Mitrovic_McWilliams_Walker_Buesing_Blundell_2021}, or by constructing counterfactual positive pairs to improve robustness under distribution shifts \citep{Roschewitz_Ribeiro_Xia_Khara_Glocker_2025}. Recent work has also studied latent causal models specifically for multimodal contrastive learning, providing theoretical links between multimodal alignment and the recovery of shared latent factors \citep{Liu_Zhang_Gong_Gao_Huang_Gong_Hengel_Zhang_Shi_2026}. In contrast to these approaches, we neither aim to recover causal structure from observed multimodal data nor use interventions to impose causal invariances. Instead, we use \acp{scm} as structured generators of synthetic multimodal dependency patterns, providing a controllable source of shared, pairwise, private, linear, and nonlinear interactions on which a multimodal scoring function can be pretrained.

\paragraph{Prior-Data Fitted Networks}
\Acfp{pfn} introduced the idea of training Transformers on synthetic datasets sampled from a predefined prior, enabling them to learn inference procedures that generalize across a distribution of tasks \citep{Müller_Hollmann_Pineda-Arango_Grabocka_Hutter_2022}. This principle has been successfully scaled to\eg tabular prediction with TabPFN \citep{Hollmann_Müller_Purucker_Krishnakumar_Körfer_Hoo_Schirrmeister_Hutter_2025}, and to\eg causal inference with Do-PFN \citep{Robertson_Reuter_Guo_Hollmann_Hutter_Schölkopf_2025} and CausalFM  \citep{Ma_Frauen_Javurek_Feuerriegel_2026}. Beyond tabular inference, CauKer demonstrates that causally structured synthetic data can also provide effective pretraining distributions for time-series foundation models \citep{Xie_Feofanov_Zhang_Palpanas_Redko_2026}. We adopt the central idea of fitting a neural model to a synthetic prior, but distinguish from existing \acp{pfn} in both objective and use: rather than learning an in-context predictor or causal estimator, we prior-fit a Transformer that serves directly as the compatibility function of a multimodal contrastive objective. The resulting scorer is subsequently frozen and transferred to real-world multimodal data.

\section{Method}
\label{sec:method}
Most contrastive objectives define multimodal similarity through a manually chosen analytical score, such as cosine similarity or a particular higher-order geometric construction. Yet there is, in principle, no reason why a specific function should be defined a priori: if one handcrafted interaction is useful, another may be as well. This motivates a more general view in which the scorer is itself treated as a nonlinear function over the modality embeddings. 
We denote a multimodal dataset $\mathcal{D}$ with \(N\) samples and \(M\) modalities by
\begin{equation}
    \mathcal{D}
    =
    \left\{
    \left(
    \mathbf{x}^{(n)}_m
    \right)_{m=1}^{M}
    \right\}_{n=1}^{N},
\end{equation}
where \(\mathbf{x}^{(n)}_m\) denotes modality \(m\in\{1,\ldots,M\}\) for sample \(n\).
Each modality is mapped into a common embedding space by a modality-specific encoder \(E_m\),
\begin{equation}
    \mathbf{e}^{(n)}_m
    =
    E_m\!\left(\mathbf{x}^{(n)}_m\right)
    \in
    \mathbb{R}^{d_{\mathrm{in}}}.
\end{equation}

\paragraph{Transformer Scorer}
Given the resulting modality embeddings
\(
\{\mathbf{e}^{(n)}_m\}_{m=1}^{M}
\),
we define their similarity through a Transformer-based scoring function \(s_{\theta}\).
The modality embeddings are processed together with a learnable CLS token, and the resulting CLS representation \(\mathbf{h}^{(n)}_{\mathrm{CLS}}\) is mapped to a scalar score,
\begin{equation}
    s_{\theta}
    \left(
        \mathbf{e}^{(n)}_1,\ldots,\mathbf{e}^{(n)}_M
    \right)
    =
    \mathbf{w}^{\top}
    \mathbf{h}^{(n)}_{\mathrm{CLS}}
    +
    b.
\end{equation}
Here, \(\theta\) denotes the parameters of the Transformer and its modality-specific input projections, while \(\mathbf{w}\) and \(b\) parameterize the final linear readout. In contrast to prescribing a fixed manual similarity function, this formulation allows the scorer to represent a broad class of nonlinear and higher-order interactions between modality embeddings.
This formulation replaces a manually specified similarity rule with a flexible function class, but leaves open which particular scoring function should be used. In the following, we address this question by shaping the Transformer scorer through prior-fitting on synthetic multimodal data generated from \acp{scm}.

\paragraph{Structural Causal Model Prior}
\begin{figure}[t]
    \centering
    \includegraphics[width=\linewidth]{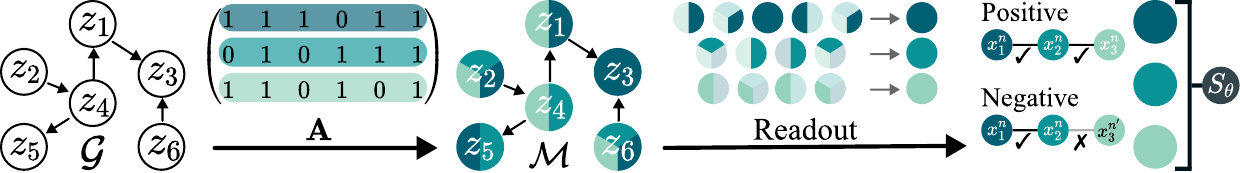}
    \caption{
        Overview of synthetic multimodal data generation and contrastive prior-fitting. A \acs{dag} $\mathcal{G}$ defines causal dependencies among latent variables, while the binary assignment matrix $\mathbf{A}$ selects the latent subsets contributing to each modality, inducing shared, subset-specific, and modality-private information. The selected latent variables are combined through linear and nonlinear readouts to produce modality vectors. During prior-fitting, matched tuples contain modalities from the same synthetic sample, whereas negatives are formed by replacing the target modality with a sampled mismatched candidate before scoring with $s_{\theta}$.
    }
    \label{fig:scm_generation}
\end{figure}
To generate samples from a multimodal \ac{scm}-based prior (\Cref{fig:scm_generation}), we sample a \ac{dag} $\mathcal{G}$ over $K$ vector-valued latent variables $\{\mathbf{z}_1,\ldots,\mathbf{z}_K\}$. We first sample a random topological ordering and then include each forward edge independently with probability $p_{\text{edge}}$. Consequently, a node may have zero, one, or multiple parents and children, while the number of root nodes is determined by the sampled graph. Each non-root latent variable is generated according to
\begin{equation}
    \mathbf{z}_j
    =
    f_j\!\left(
        \{\mathbf{z}_k : k\in\operatorname{Pa}(j)\}
    \right)
    +
    \sigma_z\boldsymbol{\epsilon}_j,
\end{equation}
where $\operatorname{Pa}(j)$ denotes the parents of node $j$, $\boldsymbol{\epsilon}_j\sim\mathcal{N}(\mathbf{0},\mathbf{I}_d)$ is
standard Gaussian noise, and $\sigma_z\geq 0$ controls its magnitude.
Root variables are sampled directly from
$\mathcal{N}(\mathbf{0},\mathbf{I}_d)$. 
The structural mechanisms\ie edges of the \ac{dag}, $f_j$ are sampled from a family of linear and nonlinear functions, allowing the prior to represent heterogeneous causal dependencies.
A binary assignment matrix
\(\mathbf{A}\in\{0,1\}^{M\times K}\)
determines which latent variables directly affect each modality\ie each latent node is assigned to modalities by $\mathbf{A}$. Both root and non-root latent variables can be selected by $\mathbf{A}$. 
The assignment matrix can create latent factors that are shared across all modalities, shared by arbitrary subsets of modalities, or private to individual modalities.
For each sample $n$, we independently draw the noise variables and evaluate the \ac{dag} to obtain $\{\mathbf{z}^{(n)}_1,\ldots,\mathbf{z}^{(n)}_K\}$. 
The \(m\)-th row of \(\mathbf{A}\) selects the complete set of latent variables used to construct modality \(m\),
\begin{equation}
    \mathcal{Z}^{(n)}_m
    =
    \left\{
        \mathbf{z}^{(n)}_j
        \,:\,
        A_{mj}=1
    \right\},
    \qquad m\in\{1,\ldots,M\}.
\end{equation}
The selected latent variables are then mapped into a modality vector $\mathbf{x}^{(n)}_m$. For notational simplicity, we suppress the sample index below.
Given the latent variables, modality $m$ is created using a mixture of linear and nonlinear readouts,
\begin{equation}
    \widetilde{\mathbf{x}}_m
    =
    \underbrace{
    (1-\alpha)
    \sum_{j=1}^{K}
    A_{mj}\mathbf{z}_j\mathbf{P}_{mj}
    }_{\text{linear readout}}
    +
    \underbrace{
    \vphantom{\displaystyle\sum_{j=1}^{K}}
    \alpha\,g_m\!\left(
    \operatorname{vec}\!\left(
    [A_{mj}\mathbf{z}_j]_{j=1}^{K}
    \right)\right)
    }_{\text{nonlinear MLP readout}}
    +
    \underbrace{
    \vphantom{\displaystyle\sum_{j=1}^{K}}
    \sigma_m\boldsymbol{\eta}_m
    }_{\text{modality-specific noise}}.
    \label{eq:modality_readout}
\end{equation}
The resulting modality vector is subsequently $\ell_2$-normalized before being passed to the scorer.
Here, $A_{mj}\in\{0,1\}$ indicates whether latent variable $\mathbf{z}_j$ contributes to modality $m$, $\mathbf{P}_{mj}\in\mathbb{R}^{d\times D}$ projects the $d$-dimensional latent variable into the $D$-dimensional modality space, and $g_m:\mathbb{R}^{Kd}\rightarrow\mathbb{R}^{D}$ is an \ac{mlp} readout applied to the ordered concatenation of the masked latent variables.
The coefficient $\alpha\in[0,1]$ controls the mixture of the linear and nonlinear readouts, while $\boldsymbol{\eta}_m\sim\mathcal{N}(\mathbf{0},\mathbf{I}_D)$ denotes standard modality-specific Gaussian noise and $\sigma_m\geq0$ controls
its magnitude. 
Across sampled \acp{scm}, we vary the graph structure, latent dimensionality, structural mechanisms, modality assignments, and noise levels. This defines a prior over multimodal dependency structures rather than a single synthetic data-generating process.

\paragraph{Prior-Fitted Scorer}
We prior-fit the Transformer scorer across multiple datasets generated from the \ac{scm} prior. 
Sampling from the \ac{scm} yields a modality vector $\mathbf{x}^{(n)}_m\in\mathbb{R}^{D}$ for each modality, where $D=d_{\mathrm{in}}$ is chosen to match the scorer input dimension.
Consequently, during prior-fitting, no encoders are required.
For each sampled \ac{scm} $\mathcal{M}$, we generate a matched \(M\)-modal dataset $\mathcal{D}_{\mathcal{M}}$. 
For a target modality \(m\), a positive tuple contains modality vectors with the same sample index \(n\), whereas negative tuples replace \(\mathbf{x}^{(n)}_m\) with \(\mathbf{x}^{(n')}_m\), \(n'\neq n\), sampled from another observation generated by the same \ac{scm}.
The Transformer outputs a scalar compatibility score for each candidate tuple. 
We use a pair-sampling approach \citep{Rheude_Hegselmann_Eils_Wild_2026} which enables contrastive learning beyond bimodal datasets with candidate-dependent scorers such as Transformers. 
For a retrieval direction, the remaining modality embeddings of an anchor are kept fixed, while the target embedding is varied over one matched candidate and sampled negatives. The scorer is evaluated independently for each resulting tuple, and a cross-entropy loss $\mathcal{L}$ encourages the matched candidate to receive the highest score. We apply this objective symmetrically over all retrieval directions.
For the frozen \ac{pfs}, the Transformer, including its input projections and scalar readout, is held fixed while the real-data modality encoders are trained to map their inputs into the scorer's input space.

\section{Experiments}
\label{sec:experiments}

\subsection{Datasets}
\label{sec:datasets}
\begin{table*}[t]
    \centering
    \caption{
        Trimodal benchmark datasets with retrieval modality as first listed modality.
    }
    \label{tab:dataset_summary}
    \small
    \begin{tabular}{@{}l c c@{}}
    \toprule
    \textbf{Dataset} & 
    \textbf{\# Samples} & 
    \textbf{Modalities} \\
    \midrule

    MIMIC-IV \citep{Johnson_Bulgarelli_Shen_Gayles_Shammout_Horng_Pollard_Hao_Moody_Gow_etal._2023,Johnson_Bulgarelli_Pollard_Horng_Celi_Mark}
        & $52,802$ & Reports, Clinicals, Demographics \\

    MIMIC-Symile \citep{Saporta_Puli_Goldstein_Ranganath_2024}
        & $10,345$ & Chest X-rays, \acsp{ecg}, laboratory \\

    \acs{ukb} \citep{Sudlow_Gallacher_Allen_Beral_Burton_Danesh_Downey_Elliott_Green_Landray_etal._2015}
        & $486,400$
        & Proteomics, Metabolomics, \acs{ehr} \\

    INSPECT \citep{Huang_Huo_Steinberg_Chiang_Lungren_Langlotz_Yeung_Shah_Fries_2023}
        & $22,449$
        & CTs, Reports, \acs{ehr} \\

    \bottomrule
\end{tabular}
\end{table*}
We focus on clinical and biomedical datasets because they provide a heterogeneous multimodal setting with images, physiological signals, tabular measurements, text-derived representations, and substantial modality missingness: 
MIMIC-IV \citep{Johnson_Bulgarelli_Pollard_Horng_Celi_Mark, Johnson_Bulgarelli_Shen_Gayles_Shammout_Horng_Pollard_Hao_Moody_Gow_etal._2023,Goldberger_Amaral_Glass_Hausdorff_Ivanov_Mark_Mietus_Moody_Peng_Stanley_2000}, 
MIMIC-Symile \citep{Saporta_Puli_Goldstein_Ranganath_2024}, 
INSPECT \citep{Huang_Huo_Steinberg_Chiang_Lungren_Langlotz_Yeung_Shah_Fries_2023}, 
and the \ac{ukb} \citep{Sudlow_Gallacher_Allen_Beral_Burton_Danesh_Downey_Elliott_Green_Landray_etal._2015} (\Cref{tab:dataset_summary}). These datasets represent different cohort sizes and clinical settings (\Cref{tab:dataset_summary}), combine raw inputs with precomputed embeddings, and include both complete and missing modality configurations. Together, they enable comparing established multimodal objectives while testing whether the proposed scorers transfer across heterogeneous datasets.

\paragraph{MIMIC-IV}
We use MIMIC-IV \citep{Johnson_Bulgarelli_Pollard_Horng_Celi_Mark,Johnson_Bulgarelli_Shen_Gayles_Shammout_Horng_Pollard_Hao_Moody_Gow_etal._2023,Goldberger_Amaral_Glass_Hausdorff_Ivanov_Mark_Mietus_Moody_Peng_Stanley_2000} with radiology reports, clinical time series, and demographic variables analogously to \cite{Rheude_Eils_Wild_2026b}. Radiology reports are represented using a pretrained RadBERT encoder \citep{Yan_McAuley_Lu_Du_Chang_Gentili_Hsu_2022}, followed by a trainable residual \ac{mlp} \citep{Touvron_Bojanowski_Caron_Cord_El-Nouby_Grave_Izacard_Joulin_Synnaeve_Verbeek_etal._2023} projection head. Clinical time series and demographic features are encoded using modality-specific residual \acp{mlp}. 

\paragraph{MIMIC-Symile}
We additionally evaluate on MIMIC-Symile \citep{Saporta_Puli_Goldstein_Ranganath_2024}, which combines chest X-rays, \acp{ecg}, and laboratory measurements as a benchmark for higher-order multimodal contrastive learning. Following the original setup, chest X-rays and \acp{ecg} are encoded using ResNet encoders \citep{He_Zhang_Ren_Sun_2016,Saporta_Puli_Goldstein_Ranganath_2024}, while laboratory measurements are represented using a residual \ac{mlp}. Each sample pairs an \ac{ecg} and laboratory measurements collected within 24 hours of admission with a chest X-ray acquired 24-72 hours after admission. Retrieving this future chest X-ray from the \ac{ecg} and laboratory values using the learned score therefore represents a zero-shot, prospective cross-modal prediction task (\Cref{fig:overview})
\citep{Saporta_Puli_Goldstein_Ranganath_2024}.

\paragraph{\ac{ukb}}
We further consider the \ac{ukb} \citep{Sudlow_Gallacher_Allen_Beral_Burton_Danesh_Downey_Elliott_Green_Landray_etal._2015}, using proteomics, metabolomics, and \acp{ehr} \citep{Rheude_Eils_Wild_2025}. The \ac{ehr} modality is represented using Qwen3-Embedding-8B embeddings \citep{Zhang_Li_Long_Zhang_Lin_Yang_Xie_Yang_Liu_Lin_etal._2025,Hegselmann_VonArnim_Rheude_Kronenberg_Sontag_Hindricks_Eils_Wild_2026}, and all three modality encoders are implemented as residual \acp{mlp}. We implement the union setting with $486{,}400$ participants, which contains substantial modality missingness and therefore reflects a realistic large-scale multimodal scenario. Retrieval evaluation is filtered for samples with the available proteomics modality.
For modality availabilities, metabolomics is almost complete with a missing rate of only \(0.87\%\). \ac{ehr} data is missing for $25.82\%$. The expensive proteomics modality is substantially sparser, with a missing rate of $89.25\%$.

\paragraph{INSPECT}
The INSPECT dataset \citep{Huang_Huo_Steinberg_Chiang_Lungren_Langlotz_Yeung_Shah_Fries_2023} comprises $22,449$ samples for \ac{pe} diagnosis, combining \ac{ct} volumes, \acp{ehr}, and radiology reports. We filter to complete multimodal samples with available binary PE labels. EHRs are represented using MOTOR embeddings \citep{Steinberg_Fries_Xu_Shah_2024}, projected with a residual \ac{mlp}, \ac{ct} volumes use spatial feature maps from the final Med3DVLM encoder stage \citep{Xin_Ates_Gong_Shao_2025}, aggregated by a lightweight 3D ResNet, and report impressions are encoded with RadBERT \citep{Yan_McAuley_Lu_Du_Chang_Gentili_Hsu_2022}, adapted using LoRA and a residual \ac{mlp}. For vision retrieval conditioned on \acp{ehr} and reports, negative candidates are drawn from different patients but restricted to the same \ac{pe} label as the query, preventing shortcut retrieval based solely on \ac{pe} status.

\subsection{Experimental Setup}

\paragraph{Training and Evaluation}
All methods are implemented in a common PyTorch Lightning framework \citep{Falcon_ThePyTorchLightningteam_2019} to ensure consistent training, optimization, encoder choices, and evaluation across methods and datasets. We perform extensive hyperparameter tuning for all methods (\Cref{app:hyperparameter_tuning}), design encoders to be geometry preserving, and use a learnable logit scale \citep{Rheude_Eils_Wild_2026a,Rheude_Eils_Wild_2026b,Rheude_Hegselmann_Eils_Wild_2026}. Models are optimized with ScheduleFree-AdamW \citep{Defazio_Yang_Khaled_Mishchenko_Mehta_Cutkosky_2024} together with norm-based gradient clipping. 
For the Transformer scorer, we do not include positional encodings because the modality tokens have no spatial or temporal ordering. Modality identity is retained through modality-specific projection networks.
Each modality embedding is first mapped to the Transformer dimension through a modality-specific projection, after which a learnable CLS token is concatenated and the resulting tokens are processed by a pre-layer-normalized Transformer encoder. 
All linear layers are initialized with Xavier initialization \citep{Glorot_Bengio_2010}, while the CLS token is initialized from \(\mathcal{N}(0,0.02^2)\).
Missing modalities are zero imputed and concatenated with binary indicators identifying which input features are missing \citep{Saporta_Puli_Goldstein_Ranganath_2024}. For the Transformer scorer missing modality tokens are excluded from self-attention using a key-padding mask. 
Results on real-world datasets are computed using 5-fold cross-validation with non-overlapping sample identifiers across folds and by reporting mean $\pm$ \ac{se}. We use a common batch size of \(128\) for all datasets. 

\paragraph{Prior-Fitting}
The \ac{scm} prior emphasizes linear structural mechanisms while introducing small nonlinear contributions through the modality-specific readouts (\Cref{app:hyperparameter_tuning}).
Latent and modality-specific noise are sampled to expose the scorer to varying noise levels and encourage robust transfer. 
Training and model selection follow two stages. First, the \ac{scm} prior-fitting hyperparameters are selected by minimizing the validation loss on held-out synthetic \acp{scm}. 
General \ac{scm} parameters\eg the structural mechanisms, number of nodes, and latent dimensionality, are fixed based on preliminary experiments to avoid overly simple synthetic tasks and potential shortcut solutions. For example, reducing the number of nodes can lower the synthetic validation loss by simplifying the prior while yielding a scorer that transfers less effectively to real-world datasets.
The resulting single \ac{pfs} is then frozen and reused for all real-world datasets, without dataset-specific prior-fitting or checkpoint selection. Second, the real-data encoders, learnable logit scale, and downstream optimization hyperparameters are tuned separately for each dataset by minimizing its validation loss. Thus, selection of the \ac{pfs} is independent of all real-world datasets. The complete search spaces and selected values are reported in
\Cref{app:hyperparameter_tuning}.

\subsection{Downstream Performance}
\begin{table}[t]
    \centering
    \caption{
        Comparison of well-tuned models optimized with different multimodal contrastive objectives. Values represent top-1 accuracy of the retrieval task (mean $\pm$ \acs{se}). Best non-overlapping value bold.
    }
    \label{tab:results}
    \small
    \begin{tabular}{@{}l c c c c@{}}
    
    \toprule
    
    \textbf{Scorer} & 
    \textbf{INSPECT $\uparrow$} &
    \textbf{\acs{ukb} $\uparrow$} &
    \textbf{MIMIC-Symile $\uparrow$} &
    \textbf{MIMIC-IV $\uparrow$} \\
    
    \midrule

    \multicolumn{5}{l}{\hspace{-6pt}\textit{Baselines}} \\
    Clip 
        & $0.5548 \pm 0.005$
        & $0.4738 \pm 0.008$
        & $0.3109 \pm 0.006$
        & $0.6503 \pm 0.001$ \\

    Triangle 
        & $0.5183 \pm 0.014$
        & $0.5259 \pm 0.009$
        & $0.4477 \pm 0.005$
        & $0.6476 \pm 0.003$ \\

    Gram 
        & $0.4586 \pm 0.012$
        & $0.4623 \pm 0.009$
        & $0.3240 \pm 0.005$
        & $0.6454 \pm 0.002$ \\

    Symile 
        & $0.6979 \pm 0.003$
        & $0.5554 \pm 0.005$
        & $0.4707 \pm 0.008$
        & $0.6562 \pm 0.003$ \\
    \midrule 

    \multicolumn{5}{l}{\hspace{-6pt}\textit{Proposed}} \\
    Transformer (frozen)
        & $0.6979 \pm 0.002$
        & $0.4462 \pm 0.005$
        & $\mathbf{0.5021 \pm 0.019}$
        & $\mathbf{0.6626 \pm 0.003}$ \\
    \ac{pfs} (frozen)
        & $\mathbf{0.7173 \pm 0.002}$
        & $\mathbf{0.5745 \pm 0.004}$
        & $\mathbf{0.5164 \pm 0.007}$
        & $\mathbf{0.6743 \pm 0.002}$ \\

    \bottomrule
\end{tabular}
\end{table}
We compare well-tuned models optimized with different multimodal contrastive objectives with a randomly initialized frozen Transformer and an \ac{scm}-prior-fitted Transformer (\ac{pfs}) across different datasets (\Cref{tab:results}).

\paragraph{Frozen Transformer}
Even without prior-fitting or downstream updates to the scorer, the frozen Transformer is competitive on three of the four datasets. 
It matches the best baseline\ie Symile, on INSPECT and further exceeds all baselines on MIMIC-Symile and MIMIC-IV. This indicates that the modality encoders can adapt effectively to a fixed nonlinear scoring function. 
However, its substantially weaker UKB performance shows that the architecture and masking mechanism alone do not provide uniformly strong transfer.

\paragraph{\ac{pfs}}
\begin{wrapfigure}[18]{r}{0.65\textwidth}
        \centering
        \vspace{-13pt}
        \includegraphics[width=\linewidth]{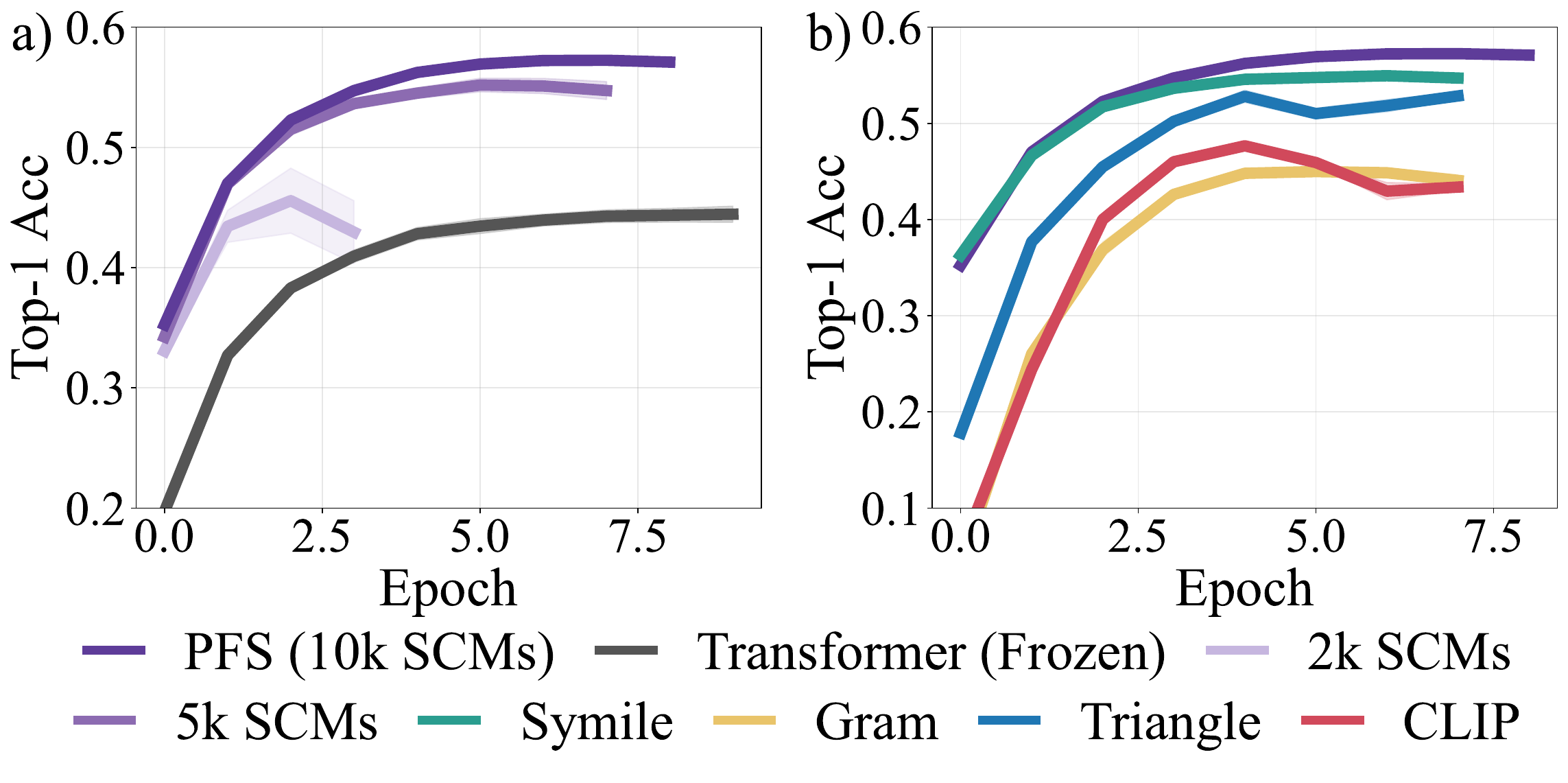}
        \caption{
            Validation performance of well-tuned baselines, the frozen Transformer and the \ac{pfs} on the \ac{ukb}.
            \textbf{a)} Scaling the prior-fitting with more sampled \acp{scm} improves transfer to real-world data.
            \textbf{b)} Using the frozen \ac{pfs} leads to faster convergence and higher retrieval accuracy than the baselines.
        }
        \label{fig:scaling_and_scorers}
\end{wrapfigure}
\ac{scm}-based prior fitting improves mean retrieval accuracy on all four datasets while also reducing variability across folds and seeds, as reflected by equal or lower \ac{se}.
The \ac{pfs} consequently achieves the highest accuracy across all benchmarks. Overall, prior-fitting is particularly valuable in settings where a frozen Transformer scorer transfers poorly, as demonstrated on the \ac{ukb}.
The \ac{pfs} benefits from more \acp{scm} during prior fitting, with performance improving as the number of sampled \acp{scm} increases from \(2{,}000\) to \(10{,}000\) (\Cref{fig:scaling_and_scorers}a). Beyond this range\eg up to \(100{,}000\) \acp{scm}, we do not observe further gains, suggesting that scaling the prior may require more architectural optimization.
In addition to higher retrieval scores, the models using the \ac{pfs} also demonstrate faster convergence compared to the baselines (\Cref{fig:scaling_and_scorers}b).

\subsection{Ablation Study}
\begin{table}[t]
\centering

\begin{minipage}[t]{0.48\textwidth}
    \centering
    \caption{
        MIMIC-Symile: Ablating the strong performance of frozen Transformers.
    }
    \label{tab:ablation}
    \small
    \begin{tabular}{@{}l c@{}}
    
    \toprule
    
    \textbf{Ablation} & 
    \textbf{Top-1 Acc $\uparrow$} \\
    \midrule

    Strongest Baseline (Symile)
        & $0.4707 \pm 0.008$ \\
    
    \midrule
    \multicolumn{2}{l}{\hspace{-6pt}\textit{Scorer Architecture}} \\
    \acs{mlp} scorer (frozen)
        & $0.2010 \pm 0.018$ \\
    Residual \acs{mlp} scorer (frozen)
        & $0.2351 \pm 0.013$\\
    Transformer scorer
        & $0.4203 \pm 0.011$ \\
    \acs{mlp} scorer
        & $0.4304 \pm 0.014$ \\
    \ac{pfs}
        & $0.4860 \pm 0.011$ \\
    Transformer scorer (frozen)
        & $0.5021 \pm 0.019$ \\
    \ac{pfs} (frozen)
        & $0.5164 \pm 0.007$ \\

    \midrule
    \multicolumn{2}{l}{\hspace{-6pt}\textit{Scorer Initializations}} \\
    Mimetic 
        & $0.2520 \pm 0.016$ \\
    Kaiming 
        & $0.3815 \pm 0.009$ \\
    Xavier 
        & $0.5021 \pm 0.019$ \\
    \ac{pfs} (frozen)
        & $0.5164 \pm 0.007$ \\

    \bottomrule
    
    \end{tabular}
\end{minipage}
\hfill
\begin{minipage}[t]{0.48\textwidth}
    \centering
    \caption{
        \ac{ukb}: Ablating the unusually weak performance of frozen Transformers.
    }
    \label{tab:ablation_ukb}
    \small
    \begin{tabular}{@{}l c@{}}
    
    \toprule
    
    \textbf{Ablation} & 
    \textbf{Top-1 Acc $\uparrow$} \\
    \midrule

    Strongest Baseline (Symile)
        & $0.5554 \pm 0.005$ \\
    
    \midrule
    \multicolumn{2}{l}{\hspace{-6pt}\textit{Scorer Architecture}} \\
    \acs{mlp} scorer (frozen)
        & $0.2969 \pm 0.074$ \\
    Residual \acs{mlp} scorer (frozen)
        & $0.3840 \pm 0.009$\\
    Transformer scorer (frozen)
        & $0.4462 \pm 0.005$ \\
    \ac{pfs}
        & $0.4589 \pm 0.029$ \\
    Transformer scorer
        & $0.4736 \pm 0.025$ \\
    \ac{pfs} (frozen)
        & $0.5745 \pm 0.004$ \\
    \acs{mlp} scorer
        & $0.6115 \pm 0.013$ \\

    \midrule
    \multicolumn{2}{l}{\hspace{-6pt}\textit{Scorer Initializations}} \\
    Mimetic 
        & $0.3652 \pm 0.006$ \\
    Xavier 
        & $0.4462 \pm 0.005$ \\
    Kaiming 
        & $0.4998 \pm 0.004$ \\
    \ac{pfs} (frozen)
        & $0.5745 \pm 0.004$ \\

    \bottomrule
    
    \end{tabular}
\end{minipage}

\end{table}
We demonstrate scorer and prior-fitting ablations. Hyperparameters of every ablation variant are retuned. (\Cref{app:hyperparameter_tuning}).
We ablate the performance of the frozen Transformer scorer on MIMIC-Symile and the \ac{ukb} to get insights for the strong and unusually weak performance respectively (\Cref{tab:ablation,tab:ablation_ukb}).
For the \ac{pfs}, we ablate on the \ac{ukb} (\Cref{tab:ablation_pfs}). Here, prior-fitting demonstrates its largest improvement, making the \ac{ukb} better suited to distinguish between differnt \acp{scm}.

\paragraph{Scorer Ablations}
We compare scorer architectures and initialization schemes on MIMIC-Symile and the \ac{ukb} (\Cref{tab:ablation,tab:ablation_ukb}). Varying the scorer architecture ablates whether the benefit is specific to the Transformer or instead reflects more general properties of neural scorers. 
Likewise, because a frozen scorer is entirely determined by its weights, comparing initialization schemes is interesting since \acp{pfs} can be interpreted as a data-driven, prior-informed initialization of the weights.
On MIMIC-Symile, residual and standard \ac{mlp} variants perform substantially worse than the frozen Transformer, and making either the Transformer or \ac{mlp} trainable does not improve retrieval. The frozen Transformer is highly sensitive to initialization, with Xavier \citep{Glorot_Bengio_2010} outperforming Kaiming \citep{He_Zhang_Ren_Sun_2015} and mimetic initialization \citep{Trockman_Kolter_2023}, while \ac{pfs} yields the strongest result.
On the \ac{ukb}, the overall picture stays the same: we observe prior-fitting benefits and optimization difficulties when training the scorer as well.
However, the trainable \ac{mlp} clearly benefits from end-to-end optimization and reaches the highest accuracy. We do not observe this behavior on the other datasets, suggesting that the \ac{ukb} data can be learned with a comparatively simple trainable scorer which is unusually effective. This is consistent with prior work indicating that \ac{ukb} shows different multimodal behavior from MIMIC-Symile \citep{Rheude_Eils_Wild_2026b}.

\paragraph{Prior-Fitting Ablations}
\begin{wraptable}[13]{r}{0.5\textwidth}
    \centering
    \vspace{-12pt}
    \caption{
        Ablation study for the \ac{scm} design and prior-fitting the \ac{pfs} on the \ac{ukb} dataset. 
    }
    \label{tab:ablation_pfs}
    \small
    \begin{tabular}{@{}l c@{}}
        \toprule
        \textbf{Ablation} &
        \textbf{Top-1 Acc $\uparrow$} \\
        
        \midrule
        Only private information \acp{scm}
            & $0.3170 \pm 0.007$ \\
            
        Transformer scorer (frozen, Xavier)
            & $0.4462 \pm 0.005$ \\

        Permute BC
            & $0.4643 \pm 0.003$ \\
            
        2k \acp{scm}
            & $0.4723 \pm 0.015$ \\ 

        Remove \ac{dag} edges
            & $0.5431 \pm 0.008$ \\

        5k \acp{scm}
            & $0.5527 \pm 0.003$ \\

        Only linear \acp{scm}
            & $0.5733 \pm 0.003$ \\

        \ac{pfs} (10k shared/pairwise \acp{scm})
            & $0.5745 \pm 0.004$ \\            
        
        \bottomrule
    
    \end{tabular}
\end{wraptable}
Although \ac{ukb} is atypical with respect to scorer trainability, it is highly discriminative for studying prior-fitting: \ac{pfs} improves top-1 accuracy over the frozen Transformer by $12.83$ percentage points. This large gap provides substantially more resolution for distinguishing individual \ac{scm} prior components than datasets on which random and prior-fitted Transformer scorers perform similarly.
With the private-information ablation, each latent variable is assigned directly to only one modality. Although this removes direct latent overlap, \ac{dag} edges can still induce indirect dependencies between modality-specific latent variables. This reduces accuracy to $0.3170$, confirming that prior-fitting requires cross-modal dependencies. 
Randomly shuffling the sample alignment between modalities during prior fitting limits performance to $0.4643$, only slightly above the frozen Transformer, highlighting the importance of preserving correct cross-modal correspondences in the synthetic data.
Performance also scales with the diversity of the prior-fitting distribution: increasing the number of sampled \acp{scm} from $2{,}000$ to $5{,}000$ and $10{,}000$ improves accuracy (\Cref{fig:scaling_and_scorers}).
Removing all \ac{dag} edges makes every latent variable a root sampled independently from $\mathcal{N}(\mathbf{0},\mathbf{I}_d)$. 
Nevertheless, shared and pairwise assignments in $\mathbf{A}$ still induce cross-modal dependence by assigning multiple modalities to the same latent variables. This variant still achieves $0.5431$, indicating that overlapping latent assignments alone provide a substantial transferable prior, while structural dependencies through the edges of the \ac{dag} offer an additional benefit. 
Restricting the prior to purely linear dependencies by setting $\alpha=0$ (\Cref{eq:modality_readout}) and using only linear structural mechanisms along the \ac{dag} edges yields $0.5733$, effectively matching the full \ac{pfs}. Thus, cross-modal information sharing, correct correspondence, and \ac{scm} diversity are the primary contributors, whereas nonlinear mechanisms provide no measurable benefit in this setting.

\section{Limitations, Conclusion \& Future Work}
\label{sec:conclusion_futurework}
\paragraph{Limitations} 
Transformer-based scoring incurs greater wall-clock training time than analytical scoring functions (\Cref{tab:computational_efficiency}), while the \ac{pfs} additionally requires a separate SCM prior-fitting stage.\begin{table}[t]
    \centering
    \caption{
        Wall time for the frozen Transformer, \ac{pfs}, and the baselines averaged over $10$ epochs. 
    }
    \small
    \begin{tabular}{@{}l c c c c c c c@{}}
        \toprule

        \textbf{Metric} &
        \textbf{\ac{mlp}} &
        \textbf{Transformer} &
        \textbf{\ac{pfs}} &
        \textbf{Clip} &
        \textbf{Triangle} &
        \textbf{Gram} &
        \textbf{Symile} \\

        \midrule

        Training step wall time (s) $\downarrow$
            & $0.049$
            & $0.244$
            & $0.244$
            & $0.032$
            & $0.031$
            & $0.035$
            & $0.038$ \\

        \bottomrule
    \end{tabular}
    \label{tab:computational_efficiency}
\end{table}
The effect of making a neural scorer trainable is not yet fully understood. Training the \ac{mlp} scorer improves over its frozen counterpart, whereas training the Transformer reduces performance, even with a separate scorer-specific learning rate and warm-up schedule. This suggests an unresolved interaction between scorer, initialization, and joint optimization with the modality encoders.
Our experiments are restricted to trimodal clinical and biomedical settings and retrieval-based evaluation. Consequently, generalization to more modalities and other domains remains to be established. 
Moreover, the effectiveness of prior-fitting depends on how well the synthetic \ac{scm} distribution captures relevant properties of real multimodal data. Selecting and tuning this prior remains an open problem, since prior-fitting is validated on synthetic tasks rather than directly on the downstream real-world distribution.
Finally, any medical application would require additional validation, including analyses of demographic and institutional biases, subgroup performance, robustness to distribution shifts, and potential failure modes, before clinical use could be considered.

\paragraph{Conclusion} 
We investigated whether contrastive learning must be specified through a manually designed analytical function or can instead be provided by a Transformer-based scorer. Our results establish a randomly initialized frozen Transformer as a surprisingly strong baseline: it matches or exceeds well-tuned models with manual objectives. We further shape this initial inductive bias by prior-fitting the Transformer on multimodal data generated from an \ac{scm} prior. The resulting \ac{pfs} improves over the frozen Transformer across all evaluated datasets.

\paragraph{Future Work} 
Prior-fitting remains relatively underexplored for multimodal contrastive learning, leaving considerable scope for richer and more targeted priors. Future work could incorporate broader families of structural mechanisms, modality counts, information-sharing patterns, missingness processes, and domain-informed data-generating assumptions. It could also investigate how properties of the prior determine the interactions learned by the scorer and how these properties should be selected. A major bottleneck is the computational cost of prior-fitting. For every sampled \ac{scm}, each positive tuple must be contrasted against multiple negative candidates for every target modality, requiring a separate Transformer evaluation for many candidate combinations. The cost therefore grows with the number of \acp{scm}, samples, modalities, and negatives, and becomes expensive when approximating the full contrastive candidate set. Nevertheless, our results indicate positive scaling effects, motivating work on more efficient candidate sampling.

\newpage 
\clearpage

\clearpage 

\subsection*{AI use statement}
    In this work, we used generative AI tools to implement methods and to assist with translation. We have not used generative AI tools to generate synthetic data sets, help develop theoretical models or conceptual frameworks, formulate mathematical claims, provide critical ingredients for proving mathematical claims, assist in the writing of proofs, propose or refine hypotheses, design or provide feedback on methodology or experiments, clean or reformat datasets, support qualitative or thematic data analysis, or interpret results. The remaining required disclosure tasks are not applicable to this work.
    Additionally, we used generative AI tools to create or edit software code and to edit parts of the research paper to improve readability. We did not use generative AI tools for the other activities for which disclosure is recommended.
    We have reviewed all AI-assisted work. Generative AI was used during the implementation of methods and the creation or editing of software code to support the development and refinement of the required code. All AI-assisted code and implementations were manually reviewed, executed, and checked for correctness and consistency with the intended methodology before being incorporated into the final work. Generative AI was also used to assist with translation. AI-generated or AI-assisted translations were manually reviewed and revised where necessary to ensure that the intended meaning, technical terminology, and context were preserved. Finally, generative AI was used to improve the readability of parts of the manuscript, including aspects such as phrasing, grammar, and clarity. All such edits were reviewed by the authors to ensure that they did not alter the scientific meaning, claims, interpretation, or conclusions of the work.
    We take responsibility for the final content of this work, including text, claims, or artifacts produced with the aid of generative AI.

\subsection*{Ethics statement}
    This work studies methodological aspects of multimodal contrastive learning and does not involve the collection of new human-subject data. Our real-world experiments use established biomedical datasets and are conducted in accordance with their respective access, governance, and data-use requirements. We report results only at the aggregate level and do not attempt to identify individual participants. Although our experiments involve clinically relevant data, the proposed methods are intended for methodological research and are not designed or evaluated for direct clinical decision making. As with other representation-learning methods trained on observational biomedical data, learned representations may inherit biases present in the underlying datasets. Consequently, performance reported here should not be interpreted as evidence of clinical validity or fairness across patient populations.

\subsection*{Reproducibility statement}
    We provide the information required to reproduce our experiments throughout the main paper and supplementary material. The model architectures, Transformer scoring function, \ac{scm} prior, and prior-fitting procedure are described in \Cref{sec:method}. Dataset construction and preprocessing are documented in \Cref{sec:datasets} and the corresponding appendix sections. We report optimization settings, hyperparameter search spaces, evaluation protocols, and compute resources in the appendix. All reported real-world experiments use the same cross-validation and evaluation framework across competing methods. An anonymized implementation is provided with the submission and will be made publicly available upon publication.

\subsubsection*{Acknowledgments}
    The authors acknowledge the Scientific Computing of the IT Division at the Charité Universitätsmedizin Berlin for providing computational resources that have contributed to the research results reported in this paper. This research has been conducted using the UK Biobank Resource under application number 49966.

\bibliography{iclr2027_conference}
\bibliographystyle{iclr2027_conference}

\clearpage 
\newpage 
\appendix

\section{Compute Environment}
\label{app:compute_environment}

All experiments were conducted on a shared \ac{hpc} cluster comprising the following compute resources:
\begin{enumerate}
    \item 21 Dell PowerEdge R7525 nodes, each equipped with 64 AMD EPYC Rome CPU cores, 512\,GB of RAM, and one NVIDIA A100 40\,GB GPU.
    \item 2 Dell PowerEdge XE8545 nodes, each equipped with 128 AMD EPYC Milan CPU cores, 512\,GB of RAM, four NVIDIA A100 40\,GB GPUs, and four NVIDIA A100 80\,GB GPUs connected via NVLink.
\end{enumerate}

\section{Hyperparameter Tuning}
\label{app:hyperparameter_tuning}
Hyperparameter selection is performed with Bayesian optimization using validation loss as the search objective. Unless stated otherwise, we allocate \(50\) optimization runs per method. In line with established tuning recommendations \citep{Godbole_Dahl_Gilmer_Shallue_Nado_2023}, the batch size is chosen beforehand and is not included in the search space. We repeat the hyperparameter search independently for every main experiment and ablation to avoid carrying over configurations across settings. For \ac{ukb}-Union, which is substantially more computationally demanding, we reduce the search budget to \(30\) trials and to $10\%$ of the training dataset size.
The bayesian search spaces for all methods are listed in the following (\Cref{lst:clip_triangle_gram_symile,lst:transformer_frozen,lst:pfs_pretraining}). Encoders for the datasets are fixed as explained in \Cref{sec:experiments}.

\begin{listing}[ht]
\begin{minted}
[
    framesep=2mm,
    linenos,
    fontsize=\scriptsize,
]
{yaml}
method: bayes
metric:
  name: val/min_loss
  goal: minimize
early_terminate:
  type: hyperband
  min_iter: 10

modelname.logit_scale_init:
    min: -3
    max: 0
    distribution: "uniform"
modelname.emb_dim:
    values: [128]
optimizer.lr:
    min: 0.00001
    max: 0.01
    distribution: "log_uniform_values"
optimizer.warmup_steps:
    values: [0, 200, 500, 1000]
optimizer.weight_decay:
    values: [0, 0.1, 0.01, 0.001]
\end{minted}
\caption{Baseline hyperparameters related to Clip, Triangle, Gram, and Symile.}
\label{lst:clip_triangle_gram_symile}
\end{listing}

\begin{listing}[ht]
\begin{minted}
[
    framesep=2mm,
    linenos,
    fontsize=\scriptsize,
]
{yaml}
method: bayes
metric:
  name: val/min_loss
  goal: minimize
early_terminate:
  type: hyperband
  min_iter: 10

modelname.logit_scale_init:
    min: -3
    max: 0
    distribution: "uniform"
modelname.emb_dim:
    values: [128]
modelname.transformer_params.d_model:
    values: [512]
modelname.transformer_params.nhead:
    values: [2]
modelname.transformer_params.num_layers:
    values: [5]
modelname.transformer_params.dropout:
    values: [0.0]
modelname.pretrained_load_logit_scale:
    values: [True, False]
modelname.freeze_scorer:
    values: [True]
optimizer.lr:
    min: 0.00001
    max: 0.01
    distribution: "log_uniform_values"
optimizer.warmup_steps:
    values: [0, 200, 500, 1000]
optimizer.weight_decay:
    values: [0, 0.1, 0.01, 0.001]
\end{minted}
\caption{Hyperparameters related to the frozen Transformer and \ac{pfs}.}
\label{lst:transformer_frozen}
\end{listing}

\begin{listing}[ht]
\begin{minted}
[
    framesep=2mm,
    linenos,
    fontsize=\scriptsize,
]
{yaml}
method: bayes
metric:
  name: val/min_loss
  goal: minimize
early_terminate:
  type: hyperband
  min_iter: 10

modelname.logit_scale_init:
    min: -3
    max: 0
    distribution: "uniform"
modelname.emb_dim:
    values: [128]
datamodule.samples_per_scm:
    values: [32]
datamodule.scms_per_batch:
    values: [64]
encoders.latent_noise_std:
    values: [0.0,0.1]
encoders.modality_readout:
    values: ["mlp"]
encoders.modality_readout_alpha_choices:  
    values: ["[0.0,0.1,0.2]"]
encoders.n_latents_choices:  # number of nodes 
    values: ["[16,32]"]
encoders.latent_dim_choices:
    values: ["[2,4,8]"]
encoders.modality_noise_std_choices:
    values: ["[0.0,0.1]"]
encoders.edge_probability_choices:
    values: ["[0.2,0.3,0.4]"]
encoders.modality_allocation_choices:
    values:
      - "[{shared_all: 0.5, pairwise: 0.5, private: 0.0}]"
encoders.mechanism_type_choices:
    values: ["[[linear],[linear,tanh]]"]
optimizer.lr:
    min: 0.00001
    max: 0.01
    distribution: "log_uniform_values"
optimizer.warmup_steps:
    values: [0, 200, 500, 1000]
optimizer.weight_decay:
    values: [0, 0.1, 0.01, 0.001]
\end{minted}
\caption{Hyperparameters related to pretraining the \ac{pfs} on \acp{scm}.}
\label{lst:pfs_pretraining}
\end{listing}

\end{document}